\documentclass[twoside]{article}

\usepackage[preprint]{aistats2027}
\usepackage{algorithm}
\usepackage{algorithmic}

\usepackage{newfloat}
\usepackage{listings}
\floatstyle{ruled}
\newfloat{listing}{tb}{lst}{}
\floatname{listing}{Listing}

\usepackage{url}
\usepackage{booktabs}

\usepackage{amsmath,amssymb}

\newboolean{showcomments}
\setboolean{showcomments}{true} 
\ifdefined\notodocomments
  \setboolean{showcomments}{false}
\fi
\ifthenelse{\boolean{showcomments}}
 { \newcommand{\mynote}[2]{
      \fbox{\bfseries\sffamily\scriptsize#1}
        {\small$\blacktriangleright$\textsf{\emph{#2}}$\blacktriangleleft$}}}
        { \newcommand{\mynote}[2]{}}

\usepackage{tikz}
\usetikzlibrary{arrows.meta,positioning,fit,backgrounds,decorations.pathreplacing,calligraphy,calc}

\begin{document}

\twocolumn[

\aistatstitle{Learning Tunable and Auditable Specifications for~Private~Tabular~Data}

\aistatsauthor{ Laura Plein, Alexi Turcotte}

\aistatsaddress{ CISPA Helmholtz Center for Information Security\\ \texttt{firstname.lastname@cispa.de}} ]

\begin{abstract}
Synthetic tabular data lets sensitive datasets be shared, but every synthesizer in wide use releases the same artifact: a table of rows, or a trained model that emits one.
What the generator concluded about the sensitive data is either invisible, locked inside parameters, or fixed in advance as a workload of chosen marginals, and the data holder cannot read what was learned, edit it, or reuse it.

We present a method that learns a \emph{specification} from a sensitive dataset: a context-free grammar paired with declarative constraints over value domains, fitted marginals, and cross-column rules mined from the leaves of a discriminator trained to separate real rows from synthetic ones.
The specification is human-readable, allowing reviewers to inspect what information has been extracted and declare what is already public knowledge before using it for generation. Early auditing allows one to tune the specification to meet the desired privacy constraints. 
The resulting specification can be used to generate not only one synthetic dataset but an unbounded family of them, all satisfying the same stated properties.
To our knowledge, this is the first approach to produce a synthetic dataset together with a specification that generates it. The data owner can then decide if they only want to share the synthetic data or both.

On seven datasets, we measure statistical similarity, task utility, and empirical privacy.
Models trained on our synthetic data and applied to real data match or exceed AIM, MST, and DP-CTGAN 
on all datasets, while empirical privacy risk remains low.
\end{abstract}

\section{Introduction}

Hospitals, statistical agencies, and banks hold data they cannot share.
Synthetic data offers a way around this: rather than release records about real people, the data holder releases artificial records that preserve the statistical structure of the original, so that analysts outside the trust boundary can explore hypotheses, train models, and validate pipelines as though they had the real thing.
The bargain holds only if the artificial records reveal little about the real ones (a generator fit to sensitive records can memorize them) so any such release must be judged on both the utility it retains and the exposure it carries.

What the data holder actually releases, though, is opaque.
A GAN, diffusion model, or fine-tuned LLM stores what it learned inside its weights, and the recipient sees only rows.
Marginal-based synthesizers such as PrivBayes~\cite{privbayes}, MST~\cite{mst}, and AIM~\cite{aim} are more legible, since the quantities they measure are named statistics rather than weights, but they fix their hypothesis class in advance, settling what may be learned before anyone looks at the data.
In neither case can a reviewer standing between the private data and the released synthetic dataset read off what the generator concluded, judge whether to accept it, or change it, and relationships that are public knowledge are mined from the private data anyway as there is no way to declare what is already known.

In this paper, we present an approach that generates a human-readable, grammar-based \textit{specification} for the private data, including declarative constraints over dataset features, which can then be used to generate synthetic datasets.
We begin with a grammar defining the syntax of the table, one nonterminal per column admitting values of the right type, and generate from it a first, deliberately unrealistic synthetic dataset.
A discriminator trained to tell those rows from real ones then supplies the constraints: analysis of the real data yields a value domain and a fitted marginal distribution for each column, while the decision paths of the discriminator's leaves that predict \textit{real} become rules relating columns to one another.
We target the Fandango specification language, and to generate data Fandango uses evolutionary search over constraint-satisfying grammar derivations produces a family of candidate datasets.
Since our approach produces a specification, it can be inspected before it is ever run, edited where a reviewer judges a constraint too revealing, and re-run as often as an analyst likes---each re-run drawing a fresh dataset without ever touching the private data again.

We contribute (i) a method for learning a human-readable specification---a grammar plus declarative constraints---from a sensitive tabular dataset, in which every quantity taken from the private data is written down explicitly and can be read, struck, or replaced before generation; (ii) an extension of the Fandango~\cite{fandango2025} language-based generation framework with \emph{population-level} constraints, letting a specification state properties of a whole dataset rather than of a single row; and (iii) an evaluation on seven datasets against CTGAN, DP-CTGAN, AIM, and MST covering similarity, utility, and empirical privacy risk.

We evaluate our technique on seven open-source datasets popular in the literature.
These vary in size
, features
, and domain
, with a mix of classification and regression tasks.
Our synthetic data closely tracks the original distributions (w.r.t.\ Wasserstein and Jensen-Shannon metrics), models trained on it and tested on real data match or exceed DP-CTGAN, AIM, and MST at $\epsilon=1$ on all datasets, and empirical risk under the inference, singling-out, and linkability attacks implemented by Anonymeter~\cite{giomi2023unified} stay low.
We stress at the outset that our approach carries no formal privacy guarantee: we compare against differentially private baselines at a commonly used budget as a reference point for achievable utility, not as a claim of equivalent protection.

\section{Background}
\label{sec:bg}
\vspace{-0.5cm}
\paragraph{Grammars and Grammar-Based Generation}

Context-free grammars (CFGs) are a classic formalism for specifying the syntax of structured languages (e.g., programming languages, file formats, network protocols), and they can also be used to specify the format of tabular data.
Formally, a CFG is a tuple $G=(N,\Sigma,P,S)$ of \emph{nonterminals} $N$ (abstract syntactic categories such as $\langle\mathit{expr}\rangle$), \emph{terminals} $\Sigma$ (the concrete symbols of the language, disjoint from $N$), \emph{productions} $P$ of the form $A \rightarrow \alpha$ with $A \in N$ and $\alpha \in (N \cup \Sigma)^{*}$, and a \emph{start symbol} $S \in N$.
A production specifies a one-step rewrite of $A$ (e.g., $\beta A\gamma \Rightarrow \beta \alpha \gamma$), a \emph{derivation} is a sequence of such steps from $S$ recorded as a \emph{derivation tree}, and the \emph{language} $L(G) = \{ w \in \Sigma^* \,|\, S \Rightarrow^* w \}$ is the set of derivable terminal strings.

The best-known use-case is as a recognizer, or \textit{parser}, but grammars can also be used as a \textit{producer}.
Sampling derivations, i.e., repeatedly expanding nonterminals from the start symbol, turns a grammar into a data generator that emits members of $L(G)$ by construction~\cite{purdom1972, fuzzingbook}. 
This is a distinctive kind of generative model: unlike one whose distribution is learned from data, a grammar characterizes its target set exactly, producing every well-formed string and no ill-formed one. 
The same principle also underlies grammar-constrained decoding for language models~\cite{picard2021}.

\paragraph{Language-Based Generation}

A grammar captures syntax but not \emph{context-sensitive} constraints, such as a length field matching a payload or a checksum (exemplary of the classic fuzzing setting), or distributional or other requirements on generated data.
Generating data that is not merely well-formed but semantically valid therefore requires pairing a grammar with such constraints, which is the purpose of the \textit{language-based input generation} of tools like ISLa~\cite{isla2022} and Fandango~\cite{fandango2025}.
Here, grammars exist alongside semantic constraints that inputs should satisfy, which is done through a semantics-aware input generation strategy (for ISLa, this is using SMT solvers, for the state-of-the-art tool Fandango, this is through a genetic algorithm).
Our work builds directly on Fandango, as effective synthetic tabular data requires more than just syntax.
Figure~\ref{fig:spec-example}(a) shows a sample Fandango specification, which mixes a grammar with constraints over nonterminals.

\begin{figure}[t]
\centering
\footnotesize
\textbf{(a) Initial specification: syntax only}
\begin{verbatim}
<start> ::= <age> "," <sex> "," <bmi>
<age>   ::= <non_zero><digit>
<sex>   ::= "0" | "1"
<bmi>   ::= <non_zero><digit><decimal>
\end{verbatim}
\textbf{(b) Learned specification: domains, marginals, rules}
\begin{verbatim}
<age> ::= <non_zero><digit> :=
      str(max(18, min(64,
      random.uniform(18.0, 64.0))))
<bmi> ::= <non_zero><digit><decimal> :=
      max(16.815, min(52.58,
      random.gammavariate(25.011, 1.225)))
where int(<smoker>) == 1
  and float(<bmi>) > 30.0
\end{verbatim}
\caption{Partial Fandango specifications for the \texttt{insurance} dataset.
\textbf{(a)} The hand-written initial specification fixes only column types and syntax.
\textbf{(b)} After constraint extraction, each column carries a fitted domain and marginal, and mined cross-column rules appear as \texttt{where} clauses.
Everything the method learned from the private data is visible here: a reviewer can strike the \texttt{where} clause, widen the \texttt{min}/\texttt{max} bounds, or replace a fitted marginal with a publicly known one.} 
\label{fig:spec-example}
\end{figure}

\section{Approach}
\label{sec:approach}

%
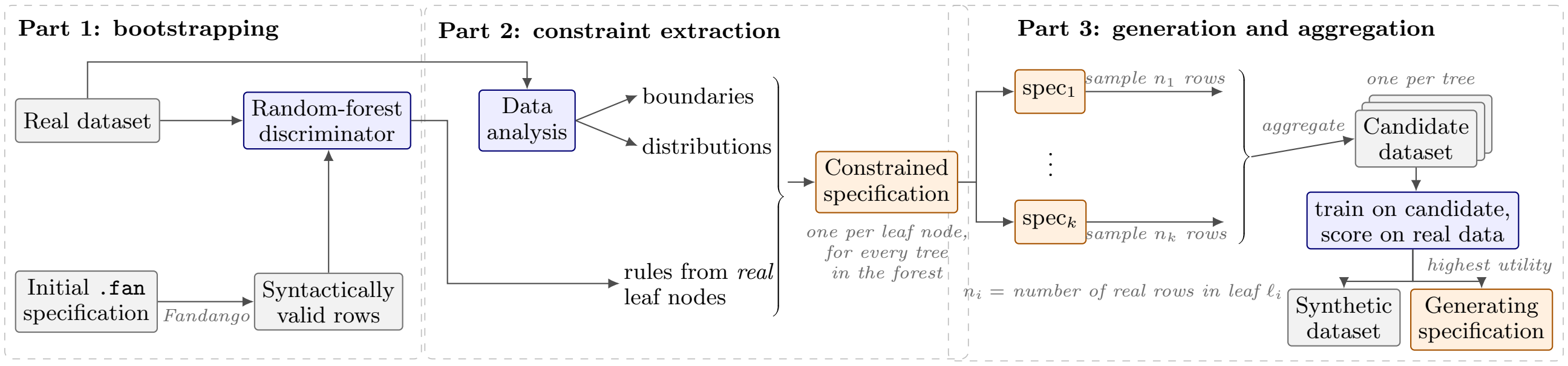
\begin{figure*}[t]
    \centering
    \resizebox{\textwidth}{!}{
\begin{tikzpicture}[
    font=\footnotesize,
    box/.style     ={draw, rounded corners=1.5pt, align=center, inner sep=3pt,
                     line width=0.5pt, minimum height=6mm},
    data/.style    ={box, fill=black!5, draw=black!55},
    proc/.style    ={box, fill=blue!7, draw=blue!45!black},
    spec/.style    ={box, fill=orange!12, draw=orange!65!black},
    prop/.style    ={align=left, inner sep=1.5pt},
    note/.style    ={align=center, font=\scriptsize\itshape, text=black!60, inner sep=1.5pt},
    ar/.style      ={-{Latex[length=4.5pt,width=4pt]}, line width=0.5pt, draw=black!70},
    brc/.style     ={decorate, decoration={calligraphic brace, amplitude=4pt},
                     line width=0.5pt, draw=black!55},
    panel/.style   ={draw=black!25, dashed, rounded corners=3pt, inner sep=4pt,
                     line width=0.4pt},
    ptitle/.style  ={font=\footnotesize\bfseries, anchor=west, inner sep=1pt},
  ]

  \node[ptitle] (t1) at ( 0.15, 4.50) {Part 1: bootstrapping};
  \node[ptitle] (t2) at ( 5.95, 4.50) {Part 2: constraint extraction};
  \node[ptitle] (t3) at (13.95, 4.50) {Part 3: generation and aggregation};

  \node[data, anchor=west] (real)  at (0.15, 3.25) {Real dataset};
  \node[proc, anchor=west] (disc)  at (3.30, 3.25) {Random-forest\\discriminator};
  \node[data, anchor=west] (ispec) at (0.15, 0.75) {Initial \texttt{.fan}\\specification};
  \node[data, anchor=west] (isyn)  at (3.45, 0.75) {Syntactically\\valid rows};

  \draw[ar] (real) -- (disc);
  \draw[ar] (ispec) -- node[note, below=1pt, pos=0.5] {Fandango} (isyn);
  \draw[ar] (isyn.north) -- (isyn.north |- disc.south);

  \node[proc, anchor=west] (an)   at (6.55, 3.25) {Data\\ analysis};
  \node[prop, anchor=west] (bnd)  at (8.75, 3.60) {boundaries};
  \node[prop, anchor=west] (dist) at (8.75, 2.90) {distributions};
  \node[prop, anchor=west] (rul)  at (8.5, 1.00) {rules from \emph{real}\\leaf nodes};
  \node[spec, anchor=west] (cspec) at (11.20, 2.40) {Constrained\\specification};

  \draw[ar] (real.north) -- ++(0,0.50) -| (an.north);
  \draw[ar] (an.east) -- (bnd.west);
  \draw[ar] (an.east) -- (dist.west);
  \draw[ar] (disc.east) -- ++(0.50,0) |- (rul.west);

  \draw[brc] (10.62, 3.85) -- (10.62, 0.55);
  \draw[ar] (10.82, 2.40) -- (cspec.west);
  \node[note, anchor=north, align=center] at ($(cspec.south)+(0,-0.10)$)
        {one per leaf node,\\for every tree\\in the forest};

  \node[spec, anchor=west] (s1) at (13.95, 3.65) {spec$_1$};
  \node                    (sd) at (14.45, 2.75) {$\vdots$};
  \node[spec, anchor=west] (sk) at (13.95, 1.85) {spec$_k$};

  \draw[ar] (cspec.east) -- ++(0.25,0) |- (s1.west);
  \draw[ar] (cspec.east) -- ++(0.25,0) |- (sk.west);

  \draw[ar] (s1.east) -- node[note, above] {sample $n_1$ rows} ++(1.90,0);
  \draw[ar] (sk.east) -- node[note, below] {sample $n_k$ rows} ++(1.90,0);

  \draw[brc] (17.05, 3.95) -- (17.05, 1.55);

  \node[data, anchor=west] (cand) at (18.65, 3) {Candidate\\dataset};
  \draw[ar] (17.22, 2.75) -- (cand.west);
  \node[note, anchor=south] at (17.93, 3) {aggregate};
  \begin{scope}[on background layer]
    \draw[data] ([shift={(0.20,0.20)}]cand.south west) rectangle
                ([shift={(0.20,0.20)}]cand.north east);
    \draw[data] ([shift={(0.10,0.10)}]cand.south west) rectangle
                ([shift={(0.10,0.10)}]cand.north east);
  \end{scope}
  \node[note, anchor=south] at (19.55, 3.65) {one per tree};

  \node[proc, anchor=north] (util) at (19.45, 2.27)
        {train on candidate,\\score on real data};
  \draw[ar] (cand.south) -- (cand.south |- util.north);

  \node[data, anchor=north] (out)  at (18.50, 0.93) {Synthetic\\dataset};
  \node[spec, anchor=north] (ospec) at (20.40, 0.93) {Generating\\specification};
  \draw[ar] (util.south) -- (19.45, 1.03) -| (out.north);
  \draw[ar] (util.south) -- (19.45, 1.03) -| (ospec.north);
  \node[note, anchor=west] at (19.58, 1.24) {highest utility};

  \node[note, anchor=north] (nn) at (15.45, 1.05)
        {$n_i =$ number of real rows in leaf $\ell_i$};

  \coordinate (p1t) at ( 1.00, 4.70);  \coordinate (p1b) at ( 1.00, 0.10);
  \coordinate (p2t) at ( 7.00, 4.70);  \coordinate (p2b) at ( 7.00, 0.10);
  \coordinate (p3t) at (15.35, 4.70);  \coordinate (p3b) at (15.35, 0.10);
  \coordinate (p3r) at (20.60, 3.40);

  \begin{scope}[on background layer]
    \node[panel, fit=(t1)(real)(ispec)(isyn)(disc)(p1t)(p1b)] {};
    \node[panel, fit=(t2)(an)(bnd)(dist)(rul)(cspec)(p2t)(p2b)] {};
    \node[panel, fit=(t3)(s1)(sk)(out)(ospec)(nn)(cand)(util)(p3t)(p3b)(p3r)] {};
  \end{scope}

\end{tikzpicture}}
    \caption{Overview of our workflow.
    \textbf{Part 1:} an initial \texttt{.fan} specification is sampled with Fandango into a syntactically valid synthetic dataset, which is paired with the real dataset to train a random-forest discriminator.
    \textbf{Part 2:} analysis of the real data yields column boundaries and marginal distributions, while the leaves of each tree that predict \emph{real} are mined for rules over feature combinations; together they form a constrained specification, one per such leaf node of every tree in the forest.
    \textbf{Part 3:} each leaf specification is sampled for as many rows as the leaf holds real rows, and the resulting populations are aggregated into one candidate dataset per tree; on each, we train a model for the dataset's downstream task (classification or regression), score it on the real data. The highest-scoring candidate is released as synthetic dataset together with the specification that generated it.}
    \label{fig:workflow}
\end{figure*}

In this section, we will describe our method for generating privacy-preserving synthetic data specifications, using the Fandango specification language and its evolutionary search-based generator. 
Our approach illustrated in Figure~\ref{fig:workflow} is composed of three main parts: 
\vspace{-0.5cm}
\begin{enumerate}
    \item Generating a first synthetic dataset from an initial specification and training a discriminator;
    \item automatically creating a constrained specification matching the distributions and feature dependencies of the original data; and 
    \item generating from partial specifications and aggregating the resulting populations into a privacy-preserving, high-quality synthetic dataset. 
\end{enumerate}
\subsection{Bootstrapping \& Discriminator Training}

For each dataset, an initial specification is written as \texttt{.fan} grammar.
The specification encodes a syntactically valid row as the only viable derivation of: 
\begin{verbatim}
<start> ::= <col_1> <delim> ... <col_n>
\end{verbatim}
For categorical columns, each category is represented as a terminal. 
For continuous numerical columns, each column is defined as a sequence of digits for integers, and two sequences of digits separated by \texttt{"."} for floats.
This simple grammar represents syntactically valid entries in the dataset, where all columns have the correct type, and we use it to produce as many samples as were part of the original dataset. 
Samples produced by this initial grammar will most likely not contain the correct feature correlations nor overall distribution, so we train a discriminator on the original and initial synthetic dataset, which tends to have very high accuracy, typically between 97\% and 99\%.

\subsection{Constrained Specification Generation}

To improve generation, we need an automated way to extract realistic \textit{boundaries}, match the overall population-level \textit{distributions}, and enforce realistic \textit{correlations} between the individual variables, and augment the specification with them.

\paragraph{Generator boundaries and distributions}

First, we perform automated data analysis on the original dataset to identify realistic boundaries and population distributions.
We start by extracting the \textit{minimum} and \textit{maximum} values for each column, ensuring that they are not isolated from the rest of the value distributions since in some datasets ``NA'' values can be replaced by, e.g., a $0$ or $-1$. 
Note: as we will be generating a specification for the private data, an auditor can freely modify these min/max values if they are concerned about them leaking private information.
To replicate the best fitting distributions, we implemented a diverse set of statistical distributions (\texttt{normal, lognormal, uniform, exponential, gamma, beta, nbinom}) for continuous variables as well as fraction distributions for categorical variables; the purpose of these is to ensure that the synthetic dataset has similar population-level metrics to the private data. 
Once these constraints are mined, the initial grammar is extended with generators that obey the constraints by construction.

\paragraph{Extending Fandango}
To be able to use Fandango with \textit{population-level} constraints, we extended the implementation of Fandango's evolutionary language-based testing framework.
There are \textit{hard} and \textit{soft} variants for constraints: hard population-level constraints \textit{must} be satisfied (within a provided error threshold), while soft constraints guide generation towards satisfying them without a strict requirement. 
This is what allows Fandango specifications to encode realistic constraints about whole datasets, not only single values.

\paragraph{Rule Extraction for Feature Correlations}

The biggest challenge in synthetic datasets is getting the cross-column constraints right.
To this end, we leverage a discriminator trained on $\frac{4}{5}$ of the original and the first synthetic dataset which confidently predicts which samples (i.e., rows) are original. The remaining $\frac{1}{5}$ will serve for the privacy evaluation.
We train a random forest classifier consisting of several decision trees of varying depth depending on the dataset size. 
We analyze each tree, focusing on leaves that predicted ``original'' samples. 
For each of those leaves, we extract the decision(s) that were taken based on what combination of conditions are present on the way to the leaf. 
Each leaf also encodes how many samples fell into it, and to avoid memorizing unique individuals that could be singled-out or inferred by later privacy attacks, we omit all leaf nodes that have fewer than five samples\footnote{
This threshold is based on preliminary experiments which identified it as the best trade-off between utility and empirical privacy.}. 
For each leaf that clears the threshold, we extract the decision conditions and convert them into constraints for \texttt{.fan} specifications, yielding the learned specification of Figure~\ref{fig:spec-example}(b).
To preserve population-level cross-column constraints, we generate partial specifications for each leaf and use them for generation. 

\paragraph{What an auditor sees}
Figure~\ref{fig:spec-example} makes the difference between the two panels exactly the set of quantities read off the private data: per-column bounds, a fitted marginal per column, and one \texttt{where} clause per retained leaf.
Each is independently reviewable and removable.
An auditor who considers a bound too tight can widen it, at the cost of some fidelity in that column; one who considers a mined rule too specific can delete it, after which the affected rows fall back to the column marginals alone.
Conversely, a relationship already in the public domain can be asserted directly and excluded from mining. 
No such edit requires re-accessing the private data; only generation is repeated.

\subsection{Sampling from Partial Specifications and Population Aggregation}

A random forest contains $t$ trees, and within each tree we retain every leaf predicting the \emph{real} class from more than four samples; writing $k$ for the number of such leaves in a given tree, this yields $k$ partial specifications for that tree, and one candidate synthetic dataset per tree.
From the partial specification for leaf $l_i$ we generate as many rows as that leaf held real samples ($n_i$ for $l_i$ in Fig.~\ref{fig:workflow}). 
In aggregating the samples from all the partial specifications of one tree, we obtain a synthetic dataset that follows the distributions of the original dataset, and we do so for each tree. 
Then, we train a model with each synthetic dataset and compare their task utility with regard to the original dataset.
Specifically, we choose the ``best'' synthetic dataset as the one which achieves the highest utility for the \textit{original dataset} when trained on it. 

Nothing in the method requires a tree ensemble as discriminator: any architecture works provided its decisions can be expressed as conditions, which for a neural discriminator means fitting a surrogate decision tree to its predictions and mining that instead.
We use a random forest because it makes this step unnecessary.

\section{Evaluation}
\label{sec:eval}

To investigate the feasibility of our novel synthetic data generation approach, we evaluate synthetic datasets across various domains in terms of \textit{similarity}, \textit{utility}, and \textit{privacy}. 
We compare our results with four state-of-the-art tools: CTGAN, DPCTGAN, AIM, and MST, and discuss the potential trade-off between utility and privacy that most techniques face.

We pose and answer the following research questions:
\begin{description}
    \item[RQ1 - Similarity] \textit{How similar in terms of statistical distributions are the original and synthetic datasets?} In this RQ, we evaluate if the statistical distributions from the original dataset are preserved in the synthetic dataset and how much distance or divergence we have for each feature.
    \item[RQ2 - Utility] \textit{Do synthetic datasets lead to the same task utility as the original datasets?} In this RQ, we train and test models on both original and generated datasets for a given task. This provides four distinct model scores to assess the overall utility of the synthetic dataset compared with the original for a given task.
    \item[RQ3 - Privacy] \textit{How high are the risks for privacy attacks on our synthetic datasets?} In this RQ, we measure the privacy risks empirically by performing several privacy attacks for data leakage and inference. We discuss the utility and privacy trade-off of our approach in comparison with the state-of-the art tabular data generation approaches.
\end{description}

\paragraph{Datasets}
For our evaluation, we use seven common publicly available datasets summarized in Table~\ref{tab:datasets}, covering classification and regression tasks over data that in a real deployment could not be released.

\begin{table}[ht]
    \centering
    \caption{Summary of datasets. Citations and links to the data can be found in the supplementary materials.}
    \begin{tabular}{llrr}
    Dataset & Task & Samples & Features \\
    \hline
    \texttt{adult}  & classification & 32562 & 15 \\
    \texttt{bank}  & classification & 4521 & 17 \\
    \texttt{cardio}  &  classification & 70000 & 13\\ 
    \texttt{diabetes}  &   classification & 768 &  9\\
    \texttt{heart}  & classification & 303 & 13 \\ 
    \texttt{insurance}  & regression & 1338 & 7 \\
    \texttt{mpg} & regression & 398 & 7 \\
    \end{tabular}
    \label{tab:datasets}
\end{table}

\paragraph{Evaluation Metrics}
Due to space limits we report a subset of results here; the rest are in the supplementary material.
We measure \textbf{utility} through the \texttt{r2-score} for regression tasks and by the \texttt{accuracy} for classification tasks.
We measure \textbf{similarity} by the Wasserstein distance~\cite{ruschendorf1985wasserstein} for continuous features and by the Jensen-Shannon~\cite{lin1991divergence} divergence for categorical features. 
For both similarity metrics, the result tables report the median of all the concerned features.
To assess the privacy of the synthetic dataset, we use Anonymeter which allows to perform various privacy attacks to measure the risks of \textit{Inference}, \textit{Singling-Out}, and \textit{Linkability} attacks.

Briefly, the three attacks are as follows.
\textbf{Inference} models an attacker who knows most attributes of a target and wishes to recover one secret attribute, guessing it from the nearest synthetic neighbor in the subspace of known attributes.
\textbf{Singling-out} tests whether an attacker can construct a predicate over attributes that matches exactly one record in the original data, either from extreme values of a single synthetic attribute (univariate) or from a conjunction over a synthetic record (multivariate).
\textbf{Linkability} tests whether two partial records drawn from disjoint attribute subsets can be tied to the same individual, by checking whether their $k$ nearest synthetic neighbors under each subset overlap.
Anonymeter always quantifies risk by comparing attacker success on the training data against a held-out control, so that only leakage attributable to the synthetic data is counted.

\subsection{Experimental Setup}
We pre-process the datasets by converting all categorical string values to integers. 
Following this conversion, we write a basic grammar for each dataset to produce syntactically valid inputs.
This step is currently manual, but mechanical: it follows directly from the column names and dtypes of a table, took a few minutes per dataset, and encodes no information beyond the schema.
Deriving it automatically is straightforward and we leave it to the implementation rather than the method. 
For the discriminator, we have chosen a Random Forest Classifier with default parameters, using the implementation from the sklearn\footnote{\url{https://scikit-learn.org/stable/modules/generated/sklearn.ensemble.RandomForestClassifier.html}} library as it allows us to easily investigate the conditions that lead to prediction decisions. 

For each dataset, we run our full pipeline, starting with initial dataset generation and analysis all the way to synthetic data generation through the candidate specifications.
The synthetic datasets are then ranked by their \texttt{gen\_orig} utility score, i.e., the utility of a model trained on the generated data when applied to the original data.
As our approach uses Fandango to generate inputs, we use the default settings, with the notable exception that we set the target number of inputs to generate to be either equal to the dataset size (in the bootstrapping phase), or to the size of the leaf (in the sub-spec generation phase). 
For the evaluation below, we only consider the top ranked synthetic dataset.

To compare our approach with the state-of-the-art, we ran CTGAN~\cite{xu2019modeling}, DP-CTGAN \cite{fang2022dp}, AIM~\cite{aim}, and MST~\cite{mst}. 
Each tool (with differential privacy, so not CTGAN) was run with varying differential privacy configurations ($\epsilon \in \{0.1, 0.5, 1, 5, 10\}$), with a fixed $\delta 
=10^{-9}$ to allow a fair comparison of results across datasets.
Since the tools provide non-deterministic results, each configuration was run five times and the best scores were considered for the utility.
Hyperparameters, per-configuration results, and the variation across candidate specifications are given in the supplementary material.

\subsection{Results}
In this section, we discuss the results in terms of \textit{similarity}, \textit{utility}, and \textit{privacy} for each dataset. 

\paragraph{RQ1 - Similarity}
To assess the similarity of the synthetic datasets with the original datasets, Table~\ref{tab:similarity} reports the median Wasserstein distance for the continuous variables and the median Jensen-Shannon divergence for categorical variables.

\begingroup
\setlength{\tabcolsep}{4pt}
\begin{table}[!h]
    \centering
    \small
    \caption{Dataset Similarity}
    \begin{tabular}{lrr}
    \toprule
        Dataset  & \multicolumn{2}{c}{Statistical Similarity}   \\
        \cmidrule(lr){2-3}
        & Wasserstein & Jensen-Shannon \\
        \midrule
        \texttt{adult}   & 0.085 & 0.253  \\
        \texttt{bank}   & 0.048 &  0.180  \\
        \texttt{cardio}   & 0.085 &  0.052 \\
        \texttt{diabetes}   & 0.099 & 0.007  \\
        \texttt{heart}   & 0.180 & 0.096   \\
        \texttt{insurance}   & 0.016 & 0.035  \\
        \texttt{mpg}   &  0.098 & 0.755\\
    \bottomrule
    \end{tabular}
    \label{tab:similarity}
\end{table}
\endgroup

Overall, the Jensen-Shannon divergence is very small, indicating a high resemblance of the statistical distributions across categorical variables. 
The divergence is only significantly higher for \texttt{mpg} where large variations for the variable \textit{year} can be observed. 
This indicates that our approach struggled to find the best fitting ratio for that distribution. 
Nevertheless, for over half of the datasets, the divergence remains below $0.1$.

\begin{figure}[!h]
    \centering
    \includegraphics[width=\linewidth]{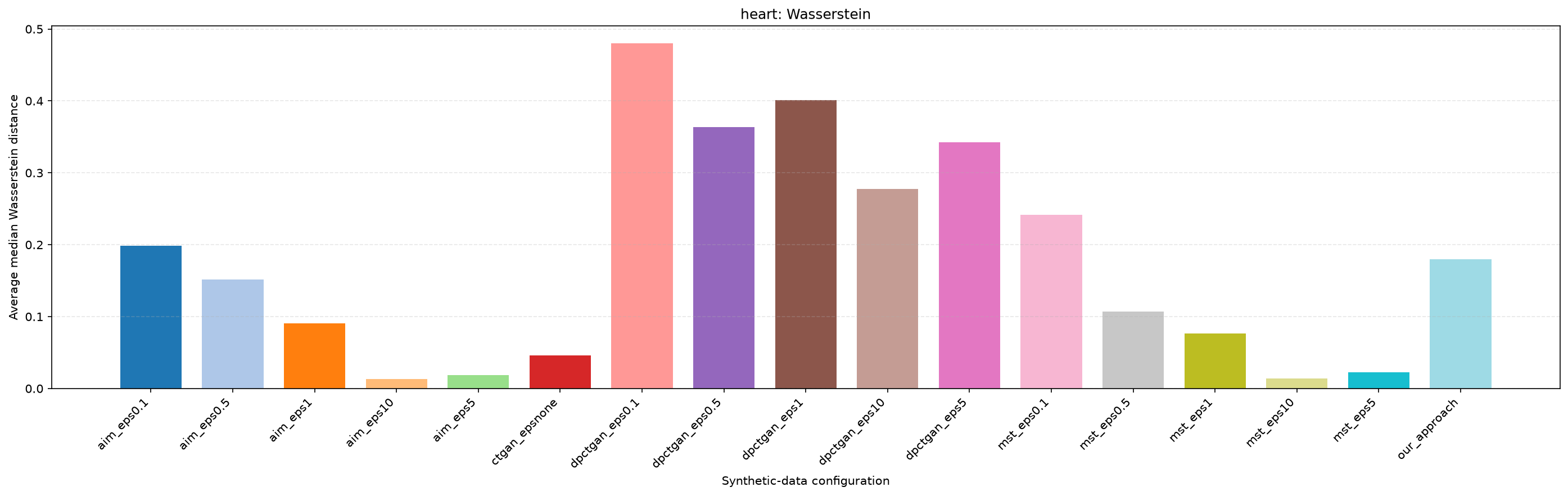}
    \caption{Wasserstein Distance for the synthetic \texttt{heart} datasets. Most right is our synthetic dataset.}
    \label{fig:wasserstein}
\end{figure}

For the Wasserstein distance, we can observe that the largest distance for continuous variables was measured on the smallest dataset. 
Here, only slight variations can significantly impact the overall distributions. 
Against the baselines, DP-CTGAN yielded larger Wasserstein distances and Jensen-Shannon divergences than ours at every budget we measured, while the marginal-based methods track the original distributions somewhat more closely---expected, since matching low-dimensional marginals is precisely what they optimize, whereas our specification reaches its distributions indirectly through constrained search.
Figure~\ref{fig:wasserstein} shows this trend, and that even our largest Wasserstein distance, on \texttt{heart}, remains small.

\begin{figure}[!h]
    \centering
    \includegraphics[width=0.45\textwidth]{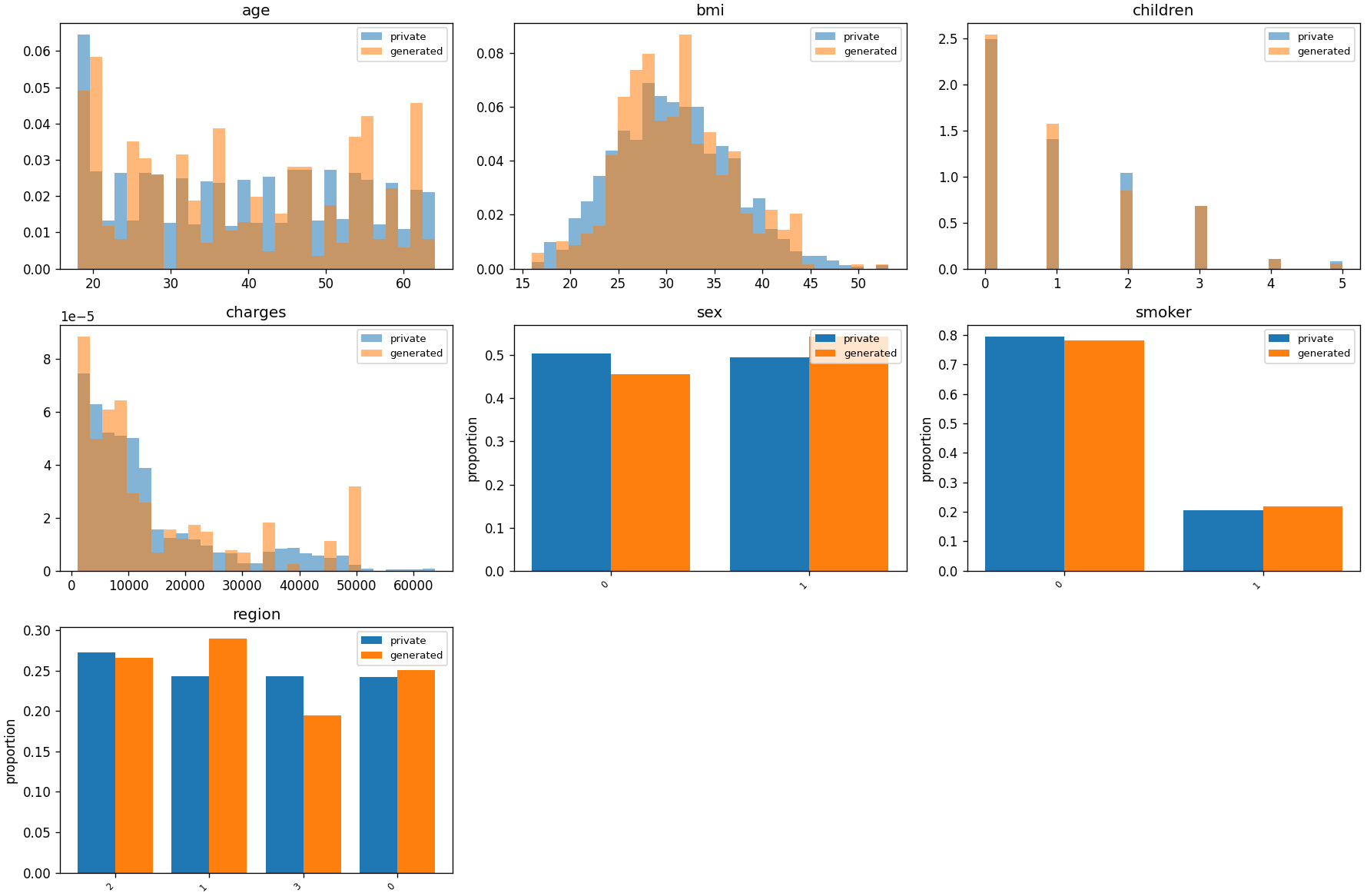}
    \caption{Statistical similarity of original and synthetic \texttt{insurance} dataset.}
    \label{fig:insurancesim}
\end{figure}

Further, Figure~\ref{fig:insurancesim} illustrates the distribution of the original and our synthetic dataset generated for \texttt{insurance}. 
While this dataset contains an almost even number of both continuous and categorical variables, it highlights our approach's capabilities to achieve the desired statistical distributions across the whole dataset. 
The plots for the remaining datasets can be found in the supplementary materials. 

\paragraph{RQ2 - Utility}
To investigate if our synthetic datasets lead to a similar task utility than the original dataset, we trained and tested two models for each dataset with a train/test split of 80/20. 
The results of the model trained and tested on original data (\texttt{orig\_orig} in Table~\ref{tab:resultsnew}) serves as upper reference score.
For each synthetic dataset, utility is reported in the right part of Table~\ref{tab:resultsnew}. 
The score for \texttt{gen\_orig} is the most relevant as it allows to train models on synthetic datasets that achieve comparable task utility while preventing or at least minimizing the risk of privacy attacks. 
Our approach achieves better or equal utility as the state-of-the-art on all datasets for the \texttt{gen\_orig} task. 
The same holds for the \texttt{orig\_gen} task where our utility is equal or reasonably close to the original, with a single exception for cardio. On \texttt{mpg} we reach only an r2-score of 0.68, as the dataset is small enough that most leaves fall below the five-sample threshold and their feature-correlation conditions are never mined---a deliberate trade of utility for privacy. 
Nevertheless, all the state-of-the-art techniques tested failed to produce a usable synthetic dataset for \texttt{mpg}.

Several baseline entries report zero utility, and the cause is the small datasets rather than a misconfiguration on our part.
For MST and AIM, the marginals measured at $\epsilon=1$ average only 1.0 rows per cell on {\tt mpg}, 13.4 on {\tt insurance}, and 19.2 on {\tt diabetes}, so the synthetic joint retains essentially no target--feature dependence.
For DP-CTGAN, sampling rate and epoch count are dataset-independent while batch size scales with sample count, so the per-step gradient signal-to-noise ratio falls from $\sim$59 on \texttt{cardio} to $\sim$1 on the smallest datasets; sweeping epochs on {\tt mpg} from 300 down to 10 raises that ratio by up to $28\times$ yet recovers no dependence at any budget, so we retain the CTGAN default of 300 throughout. Note also that the high \texttt{gen\_gen} scores (up to 1.0) are not themselves a positive result: a model that separates synthetic rows more easily than real ones indicates that the generated data is cleaner than the original, and it is the \texttt{gen\_orig} column that measures transfer to the real task.

\begingroup
\setlength{\tabcolsep}{4pt}
\begin{table}[!h]
    \centering
    \small
    \caption{Utility Results (accuracy for classification tasks and r2-score for regression tasks). Negative r2-scores are mapped to 0.00. Results from differentially private tools are reported for $\epsilon = 1$. }
    \resizebox{\columnwidth}{!}{%
    \begin{tabular}{l|r|c|rrr}
    \toprule
        Dataset & Utility & Tool & \multicolumn{3}{c}{Synthetic Data Utility}  \\
        \cmidrule(lr){4-6}
        & orig\_orig & &  gen\_gen & gen\_orig & orig\_gen \\
        \midrule
        \texttt{adult}& 0.85 & ours &  0.99 & \textbf{0.77} &  \textbf{0.82} \\
        & & CTGAN & 0.77 & 0.75 & 0.76 \\
         & & DPCTGAN & 0.91& 0.65& 0.70 \\
         & & AIM  & 0.98& \textbf{0.77} & 0.60\\
         & & MST & 1.0& 0.64& 0.71\\
        \texttt{bank}  & 0.90 & ours& 1.0 & \textbf{0.89} &  \textbf{0.93} \\
        & & CTGAN & 0.77&  0.78& 0.76\\
        & & DPCTGAN  & 0.94& 0.58& 0.72\\
        & & AIM  &0.95 &0.76 & 0.79\\
        & & MST  &0.98 & 0.81&0.80 \\
        \texttt{cardio}  & 0.73 & ours & 0.82 & \textbf{0.61} &  0.55 \\
        & & CTGAN & 0.65& 0.60& 0.61\\
        & & DPCTGAN  & 0.80& 0.56& 0.59 \\
        & & AIM  &0.94 &0.60 &0.62\\
        & & MST & 1.0& 0.56& \textbf{0.67}\\
        \texttt{diabetes}  & 0.72 &ours & 0.92 & \textbf{0.72} & \textbf{0.67}  \\
        & & CTGAN &0.09 & 0.00& 0.00\\
        & & DPCTGAN  & 0.39 & 0.00&  0.00\\
        & & AIM  & 0.20& 0.00&0.00 \\
        & & MST  & 0.94 & 0.00 & 0.00 \\
        \texttt{heart}  &0.85 &ours & 0.88  & \textbf{0.72} &  \textbf{0.64} \\ 
        & & CTGAN & 0.46& 0.53& 0.51\\
        & & DPCTGAN  & 1.0 & 0.54& 0.60\\
        & & AIM  &0.78 &0.58 &  0.55\\
        & & MST  & 0.92 &0.55 & 0.57\\
        \midrule
        \texttt{insurance}  & 0.88 & ours& 0.94 & \textbf{0.79} & \textbf{0.83} \\
        & & CTGAN &0.00 &0.00 &0.00 \\
        & & DPCTGAN  & 0.00& 0.00& 0.00\\
        & & AIM & 0.38& 0.40& 0.23\\
        & & MST  & 0.74& 0.00&0.00 \\
        \texttt{mpg} & 0.91 & ours& 0.86  & \textbf{0.68} & \textbf{0.44} \\ 
        & & CTGAN & 0.00& 0.00&0.00 \\
        & & DPCTGAN  & 0.03&0.00 &0.00 \\
        & & AIM  & 0.00& 0.00&0.00 \\
        & & MST  & 0.70& 0.00&0.00 \\
    \bottomrule
    \end{tabular}}
    \label{tab:resultsnew}
\end{table}
\endgroup

Learning a specification and sampling from it took between 6 seconds (\texttt{mpg}) and 11 minutes (\texttt{adult}).
Since leaves holding fewer than five real rows are discarded, the released datasets are consistently smaller than the originals---27\% to 46\% of the original row count---so the utility above is obtained from roughly a third of the training rows, with the omitted regions being the sparse ones that drive singling-out and linkability risk.

\paragraph{RQ3 - Privacy}
Table~\ref{tab:resultsprivacy} reports empirical risk for our synthetic datasets under each of the three attacks.
We report these for our method only: the differentially private baselines carry a formal worst-case guarantee that our approach does not, so the meaningful question for them is the budget they were run at rather than a measured attack success rate.
For reference, Giomi et al.\ evaluate these same three attacks on DP-CTGAN and find non-zero risk on all of them~\cite{giomi2023unified}.
We discuss the \textit{Inference}, \textit{Singling-Out} and \textit{Linkability} results individually.

\begingroup
\setlength{\tabcolsep}{4pt}
\begin{table*}[!h]
    \centering
    \small
    \caption{Privacy Results on our Synthetic Datasets using Anonymeter for Singling-out, Linkability and Inference attacks. }
    \begin{tabular}{lrrrrrrrr}
    \toprule
        Dataset  & \multicolumn{8}{c}{Privacy Risks}  \\
        \cmidrule(lr){2-9}
        & \multicolumn{2}{c}{Singling-Out}  & \multicolumn{3}{c}{Linkability}  & \multicolumn{3}{c}{Inference}\\
        \cmidrule(lr){2-3}\cmidrule(lr){4-6}\cmidrule(lr){7-9}
        & univariate & multivariate & $k = 1$ & $k = 5$ & $k = 10$ & min & mean & max\\
        \midrule
        \texttt{adult}   & 0.0067 & 0.0895 & 0 & 0.0004 & 0.0012 & 0 & 0.0064 & 0.0379 \\
        \texttt{bank}   & 0.0235 & 0.1400  & 0.0007 & 0.0016 & 0.0007 & 0 & 0.0062 & 0.0259 \\
        \texttt{cardio}   &  0.0191 & 0.1024 & 0 & 0.0004 & 0.0016 & 0 & 0.0051 & 0.0195 \\
        \texttt{diabetes}   & 0.0510 & 0.1185 & 0.0039 & 0.0097 & 0.0320 & 0 & 0.0074 & 0.0359 \\
        \texttt{heart}   & 0.0159 & 0.1453 & 0 & 0.0223 & 0.0602 & 0 & 0.0223 & 0.1293 \\
        \texttt{insurance}   & 0.0066 & 0.1148 & 0.0015 & 0.0088 & 0.0111 & 0 & 0.0077 & 0.0286 \\
        \texttt{mpg}   & 0.0244 & 0.0615 & 0.0099 & 0.0161 & 0.0129 & 0 & 0.0178 & 0.0683 \\
    \bottomrule
    \end{tabular}
    \label{tab:resultsprivacy}
\end{table*}
\endgroup

\paragraph{Inference} 
Across all datasets, our synthetic data resulted in an inference risk of zero for many attributes and only a slight inference risk on average. 
The highest inference risk was observed for the \texttt{heart} dataset where the ``fbs'' and ``ca'' variables were inferred the most often. 
However, ``fbs'' is a binary variable that is very unbalanced in the original dataset with a ratio of 85 to 15. 
The ``ca'' variable is also quite unbalanced across 5 categories with one being significantly more present. 
Thus, it is very likely to guess the correct value independently of our approach. 
The lowest risk was observed for \texttt{cardio}, the largest dataset in our study; inference risk decreases slightly as datasets grow, but remains low even on the small ones.

\paragraph{Singling-Out} 
One of the highest privacy risk for both attack versions was observed for the \texttt{bank} dataset. 
Since the \texttt{bank} dataset contains the most features, it is most likely easier to perform a successful multivariate attack, given the knowledge of sufficiently many variables. 
The same applies for the \texttt{heart} dataset which comes second in terms of multivariate attack risk and number of features in the dataset.

\paragraph{Linkability}
Linkability attacks are very limited with $k=1$ neighbor, only for the \texttt{mpg} dataset, the attack showed a slight impact. 
With increasing neighbors, the attacks performance increases to 0.06 in the worst case for the \texttt{heart} dataset. 
Those two datasets, being the smallest, are in line with the linkability risk, with $k=5$ being the highest on those.
This is expected: the mined rules describe correlations that hold across a population, not outlier cases, and the five-sample leaf threshold removes the narrow rules that would help an attacker bridge two attribute subsets.

\paragraph{Discussion} 

Taken together, the measured risks are low, and concentrated on the datasets with the most features, where a multivariate attacker
has the most information, and on the smallest
datasets, where each retained leaf covers a narrower slice of the population.
As noted, our approach offers no formal privacy guarantee.
Differential privacy caps the influence any single record can have on what is released, in the worst case over all datasets and adversaries, and is a strictly stronger statement than any measurement we can make; what it does \emph{not} provide is a certificate that concrete attacks fail, as the DP-CTGAN results above illustrate.
Formal guarantees and empirical risk are complementary axes, and we see the specification as a natural place to \emph{attach} differential privacy in future work rather than as an alternative to it.

\section{Related Work}
\label{sec:related}

\paragraph{Generating Synthetic Tabular Data}

In this work, we focused our evaluation against four synthetic tabular data generation techniques. 
CTGAN is a classic generative-adversarial approach, pitting a generator model against a discriminator model, training and tuning both in an attempt to get one to fool the other~\cite{xu2019modeling}.
CTGAN does not ensure the privacy of the original data, which DP-CTGAN tackles as a differentially private extension of CTGAN for synthetic medical tabular data, in which privacy is enforced via DP-SGD~\cite{Abadi_2016}, clipping and noising per-sample gradients during training to provide a formal ($\epsilon$, $\delta$)-DP guarantee.
MST is an approach that selects a set of low-dimensional marginals connected by a maximum spanning tree of mutual information, measures them under the Gaussian mechanism, and synthesizes data via the Private-PGM graphical model inference framework~\cite{mst}.
Finally, AIM adaptively selects which marginal queries to measure based on the current model's worst-approximated statistics, answering each with calibrated Gaussian noise before updating a Private-PGM model, distributing the privacy budget where it is most needed~\cite{aim}.
There are many other techniques for synthetic data generation, e.g., PrivBayes~\cite{privbayes} which skirts around the challenge of high-dimensional data by injecting noise in low-dimension marginals.

\paragraph{Learning Grammars and Constraints}

A separate line of work learns specifications from artifacts rather than using them to generate.
Grammar inference recovers a CFG describing a program's input language, either from samples alone as in GLADE~\cite{glade2017} and Arvada~\cite{arvada2021}, or from observed control flow as in Mimid~\cite{mimid2020}; dynamic invariant detection, exemplified by Daikon~\cite{daikon2001}, mines likely predicates over program state.
Both produce human-readable artifacts, and both inform our treatment of the specification as the object of interest.
The difference is the target: these techniques specify a \emph{program} in order to test it, whereas we specify a \emph{private dataset} in order to release a generator for it, making what the specification discloses a first-class concern.

\paragraph{Privacy of Generated Data}

Many synthetic data generation approaches (like those discussed above) rely on theoretical privacy guarantees afforded by differential privacy, which essentially limits the influence of any single data point on models trained from a data set.
Nevertheless, it is still possible to leak information from these datasets, and there has been some work on assessing privacy risks in generated datasets.
For example, recent work presents a framework for privacy and utility evaluation of synthetic data~\cite{Lautrup_2024}, while other work provide a toolbox for \textit{adversarial} privacy auditing of synthetic data~\cite{houssiau2022tapastoolboxadversarialprivacy}.
We used the Anonymeter~\cite{giomi2023unified} tool to assess the privacy of the synthetic datasets empirically, which automatically performs privacy attacks and quantifies the information leak risk.

\section{Conclusion}

We presented a technique that learns a human-readable specification from private tabular data and generates synthetic data from it.
The specification encodes column boundaries, population-level distributions, and cross-column constraints declaratively, so an auditor can see exactly what was learned and adjust it before generation---in contrast to techniques that release opaque models or fixed tables.
On seven datasets we achieve competitive utility at low empirical privacy risk, and we see promise in specification-based generators \textit{beyond tabular data}.

\bibliography{references}

\clearpage
\appendix
\thispagestyle{empty}



\end{document}